\documentclass[sigconf, nonacm, balance=false]{acmart}
\AtBeginDocument{%
  }

\setcopyright{none}              

\acmDOI{}
\acmISBN{}
\acmBooktitle{}
\acmYear{2026}
\acmMonth{11}
\copyrightyear{2026}
\renewcommand\footnotetextcopyrightpermission[1]{}

\usepackage{graphicx}
\graphicspath{{figures/}}
\usepackage{tabularx}
\usepackage{array}
\usepackage{placeins} 
\usepackage{ragged2e}
\usepackage{microtype}
\usepackage{algorithm}
\usepackage{algpseudocode}
\usepackage{enumitem}
\usepackage{makecell}
\usepackage{multirow}

\newcolumntype{Y}{>{\centering\arraybackslash}X}

\begin{document}

\title{ReTAMamba: Reliability-Aware Temporal Aggregation with Mamba for Irregular Clinical Time Series Prediction}


\author{Jinwoong Kim}
\orcid{0009-0001-0707-5908}
\affiliation{%
  \department{Department of Industrial Data Engineering}
  \institution{Hanyang University}
  \city{Seoul}
  \country{Republic of Korea}
}
\email{dnddl9456@hanyang.ac.kr}

\author{Sangjin Park}
\authornote{Corresponding author.}
\orcid{0009-0008-6922-624X}
\affiliation{%
  \department{Department of Industrial Data Engineering}
  \institution{Hanyang University}
  \city{Seoul}
  \country{Republic of Korea}
}
\email{psj3493@hanyang.ac.kr}


\begin{abstract}
Clinical time-series data are difficult to model with methods designed for regular sequences because they exhibit irregular sampling, frequent missing values, and heterogeneous observation patterns across variables. Existing approaches commonly use observation masks and time-gap information, but they do not continuously capture the decaying reliability of past observations or consistently organize multi-resolution information within a coherent temporal context during aggregation. To address these limitations, we propose Reliability-aware Temporal Aggregation with Mamba (ReTAMamba), which reconstructs clinical time series as time-variable token sequences, estimates observation reliability from missingness and elapsed time, and augments interval summaries with statistical descriptors. Chronological Weaving is used to integrate short- and long-term temporal information within a coherent temporal context, and a budgeted token router is applied to constrain sequence length while preserving informative summaries. Experiments on MIMIC-IV, eICU, and PhysioNet 2012 show that ReTAMamba consistently improves AUPRC over strong baselines, with average relative gains of 7.51\%, 7.80\%, and 10.15\%, respectively. Cohort-level and patient-level analyses on eICU further showed that the learned mean decay for more dynamic signals, such as heart rate and blood pressure, was 24.3\% larger than that for relatively static signals, such as laboratory test variables. These findings suggest that effective prediction in irregular clinical time series requires modeling not only what was measured, but also when and how it was observed, including information freshness and observation timeliness.
\end{abstract}




\maketitle
\makeatletter
\fancypagestyle{plain}{%
  \fancyhead[RO]{\@shortauthors}%
  \fancyhead[RE]{\@shortauthors}
  \renewcommand{\headrulewidth}{0pt}%
}
\makeatother

\pagestyle{plain} 

\section{Introduction}

Clinical time-series data collected through the widespread adoption of electronic health record (EHR) systems have become a key source for characterizing changes in patient physiological status and play an important role in severity assessment, risk prediction, and early clinical decision support \cite{ref1,ref2}. This is especially critical in intensive care units (ICUs), where patient conditions can change rapidly, making accurate prediction of outcomes such as in-hospital mortality within the first 24--48 hours of admission an important task \cite{ref3}.

Unlike regular time series, clinical time series are inherently irregular and sparse: measurement frequencies differ across variables, only a subset of variables may be observed at a given time, and intervals between measurements are highly uneven \cite{ref4}. Moreover, missingness is not merely the absence of information, but can reflect the observation process and clinical decision-making, thereby functioning as informative missingness \cite{ref5}. Clinical time-series prediction therefore requires modeling not only observed values, but also the observation structure, including which variables were measured and when, together with the freshness of information as time elapses after the last observation.

Accordingly, prior research has moved toward jointly modeling irregular observation structures, the meaning of missingness, and temporal context across multiple resolutions. Early approaches mainly regularized data through hourly aggregation or forward imputation and then applied general-purpose models, but such methods failed to preserve the clinical context embedded in the original observation patterns \cite{ref6}. Later studies attempted to model irregular sampling more directly, either by interpolating irregular observations at reference time points or by unfolding them into event sequences \cite{ref7,ref8}. However, these methods struggled to simultaneously handle cross-variable heterogeneity in observation frequency, the computational burden of long-sequence processing, and the complexity of clinical observation structures \cite{ref9}. Because missingness in clinical time series often functions as informative missingness, recent approaches have increasingly emphasized both the meaning of missingness itself and the validity of information as time elapses \cite{ref5}. In addition, accurate assessment of patient status requires temporal aggregation and multi-scale modeling that capture both short-term physiological changes and long-term disease trajectories \cite{ref10}. Nevertheless, these studies did not sufficiently resolve how to incorporate the reliability of observational signals consistently throughout representation learning or how to align information generated at different temporal resolutions along the true temporal axis while efficiently controlling the resulting increase in sequence length \cite{ref10}.

To address these challenges, this paper proposes Reliability-aware Temporal Aggregation with Mamba (ReTAMamba), a unified framework for modeling irregular clinical time series as a reliability-aware multi-scale sequence. ReTAMamba represents irregular multivariate records as time-variable token sequences, preserving variable-specific observation intervals and missingness patterns without collapsing them into conventional aggregation-based inputs. It then estimates time-varying observation reliability from missingness and elapsed time, incorporates this reliability into multi-resolution temporal aggregation, and reorders summary tokens from different temporal resolutions through Chronological Weaving. Finally, it applies budgeted token routing, using soft routing during training and hard top-$k$ selection during inference, before Mamba encoding. Through this design, ReTAMamba jointly models observation structure, information reliability, recency, multi-scale temporal context, and budgeted sequence compression within a single predictive framework.

The main contributions are as follows:
\begin{itemize}
    \item Proposing a unified token-sequence framework for irregular clinical time series that more directly preserves sparse observation structure and variable-specific missingness than conventional aggregation-based representations.
    \item Introducing a reliability-aware temporal aggregation mechanism that continuously estimates observation validity from missingness and elapsed time under variable-specific decay and incorporates it into multi-resolution summary construction.
    \item Developing a multi-scale sequence modeling strategy centered on Chronological Weaving, which reorders interval summaries from different temporal resolutions in a time-ordered manner and incorporates budgeted token routing, using soft routing during training and hard top-$k$ selection during inference, before sequence encoding.
\end{itemize}

The remainder of this paper is organized as follows: Section 2 reviews related work, Section 3 describes the proposed ReTAMamba architecture, Section 4 presents the experimental setup and results, and Section 5 concludes the paper.

\section{Related Work}

\subsection{Irregular Clinical Time-Series Modeling}

Early studies on irregularly sampled time series mainly regularized data through hourly aggregation or forward imputation, but such approaches were criticized for discarding meaningful clinical context embedded in observation patterns \cite{ref6}. Later work attempted to learn sampling irregularity directly by modifying RNN state updates or introducing time-aware attention \cite{ref5,ref11}. However, these methods still struggled to address large cross-variable differences in observation frequency, known as inter-series discrepancy \cite{ref9}. More recent approaches have followed two directions: interpolation-based methods, such as IP-Nets \cite{ref7} and mTAN \cite{ref11}, which map irregular observations to predefined reference times, and unfolding-based methods, such as SeFT \cite{ref9}, which directly represent data as $(\text{value}, \text{variable}, \text{time})$ tuple sequences. Interpolation-based methods forcibly merged information at fixed reference points, limiting their ability to jointly capture fine- and coarse-grained temporal patterns \cite{ref10}. Unfolding-based methods were shown to preserve raw observation patterns more faithfully, but in ICUs, where dozens of physiological variables are monitored over long periods, it was observed that sequence length grows rapidly with the number of observations, causing substantial memory cost and computational bottlenecks in Transformer-based architectures with $O(L^2)$ complexity and hindering real-time clinical decision support \cite{ref12}. Recently, linear-complexity state space models such as Mamba \cite{ref13} emerged as promising alternatives for large-scale sequence modeling, but effectively integrating them with clinical observation structures characterized by extreme imbalance and missingness remains an important challenge.

\subsection{Reliability-Aware Modeling}

Unlike regular time series, clinical time series exhibit variable-specific measurement intervals, and missingness itself can function as informative missingness shaped by clinical decision-making \cite{ref14,ref15}. Early approaches, such as GRU-D \cite{ref5}, model information decay using observation masks and time gaps, while more recent methods, such as Raindrop \cite{ref8}, use graph structures to learn interactions among asynchronous observations. However, these methods still struggle to address frequency differences across observational signals and their time-varying value in a structurally consistent manner \cite{ref10}. In practice, rapidly changing variables such as heart rate or respiratory rate have fundamentally different information half-lives from blood test variables that evolve on a daily scale \cite{ref16}. Consequently, although effective for short-term missing-value correction, these methods may still distort information by failing to preserve observation validity consistently throughout deeper representation learning \cite{ref10}. This calls for a framework that quantifies the exponentially decaying validity of past observations in terms of a variable-specific continuous reliability. Dynamically controlling uncertain missing information and integrating such reliability-based observation structures throughout large-scale sequence modeling remains an important challenge. At the same time, the need to jointly capture rapid, short-term physiological changes and long-term disease trajectories has highlighted the importance of multi-scale analysis in clinical time-series modeling \cite{ref10}.

\begin{figure*}[t]
    \centering
    \includegraphics[width=0.65\textwidth]{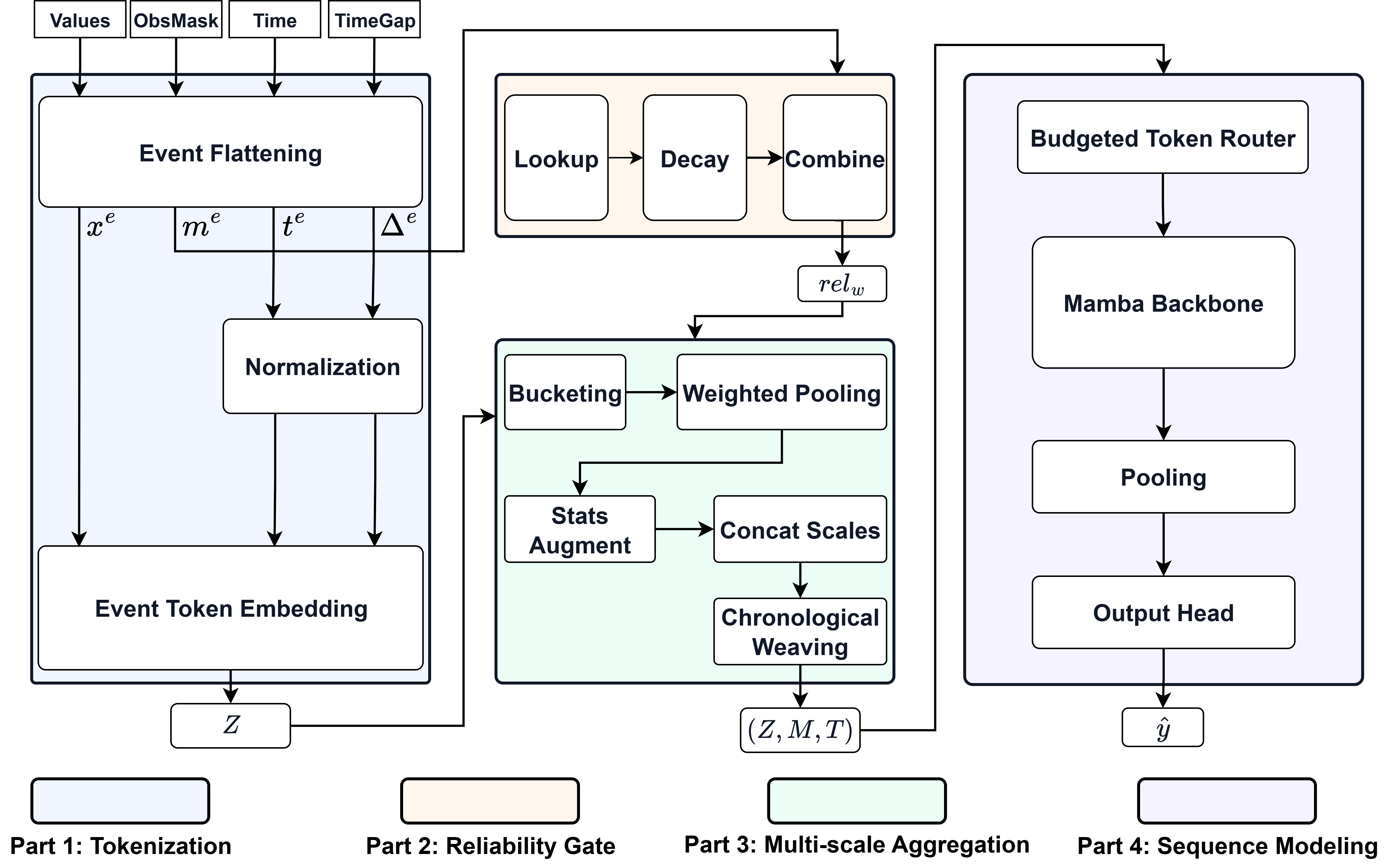}
    \caption{Detailed architecture of ReTAMamba.}
    \label{fig:retamamba_architecture}
\end{figure*}

\subsection{Multi-scale Modeling}

In EHR research, capturing patient status across multiple temporal resolutions, from minute-level physiological changes to multi-day disease trajectories, is critical for accurate prognosis prediction \cite{ref10}. Recent efforts have incorporated multi-scale mechanisms into general-purpose models such as Transformers, but these models are mainly designed for regularly sampled data and remain inefficient in sparse, irregular clinical settings \cite{ref9,ref12}. Multi-scale architectures tailored to clinical irregularity, including Warpformer \cite{ref10}, have also been proposed, yet they still construct resolution-specific summaries independently or combine them in parallel. Moreover, clinical signals inherently exhibit temporal misalignment due to differences in sampling rates and measurement precision \cite{ref8,ref17}. DTW-based alignment has been studied to address this issue, but most methods are limited to retrospective matching between regular time series and are therefore unsuitable for causal modeling with real-time inference \cite{ref18,ref19}. Supporting multi-resolution learning in irregular clinical time series therefore requires more than parallel multi-scale summarization. It requires a framework that can continuously account for time-varying reliability, organize cross-scale summaries along the temporal axis, and efficiently control sequence length. Related studies have also explored complementary directions such as efficient recurrent architectures, self-supervised representation learning, and alternative sequence representations \cite{ref30, ref31, ref32}, further highlighting the need for scalable and structure-aware learning under irregular observation patterns.

\section{Proposed Model}

Figure~\ref{fig:retamamba_architecture} illustrates the overall architecture of ReTAMamba. Each input sample consists of values $X \in \mathbb{R}^{L \times V}$, an observation mask $m \in \{0,1\}^{L \times V}$, measurement times $t \in \mathbb{R}^{L}$, and time gaps $\Delta \in \mathbb{R}_{\geq 0}^{L \times V}$, where $\Delta$ denotes the elapsed time since the last observation of each variable. Here, $L$ and $V$ denote the sequence length and number of variables, respectively. ReTAMamba transforms this irregular multivariate time series into a reliability-aware multi-scale token sequence for prediction. It first reconstructs the input as a time-variable token sequence aligned with the observation structure, then aggregates tokenized observations across multiple temporal resolutions under continuously estimated observation reliability, and finally compresses the resulting summaries under a fixed budget before sequence encoding. Through this process, ReTAMamba jointly models observation structure, information reliability, recency, and cross-scale temporal context within a single predictive pipeline.

\subsection{Tokenization}

The Tokenization stage converts an irregular multivariate time series into a time-variable token sequence that preserves variable-wise observation structure. Rather than collapsing the input into conventional interval-level summaries at the tokenization stage, it represents each time-variable cell on the shared temporal axis as a token unit, allowing sparse observations and variable-specific missingness patterns to be retained explicitly in the sequence representation. Tokenization consists of three steps: Event Flattening, Normalization, and Event Token Embedding.

\textbf{Event Flattening.} Tokenization does not aggregate variables into a single time-step representation. Instead, each time-variable cell is treated as a separate token unit and rearranged into a one-dimensional sequence aligned with the time-variable structure. The total number of token units generated from one sample is defined as follows:
\begin{equation}
N = L \cdot V,
\label{eq:1}
\end{equation}
where $N$ denotes the total number of time-variable token units generated from one sample. The observed values, observation mask, and variable-wise time gaps are flattened into one-dimensional sequences:
\begin{equation}
x^e = \mathrm{reshape}(X), \quad
m^e = \mathrm{reshape}(m), \quad
\Delta^e = \mathrm{reshape}(\Delta),
\label{eq:2}
\end{equation}
where $x^e$, $m^e$, and $\Delta^e$ are one-dimensional rearrangements of the value matrix $X$, observation mask $m$, and variable-wise time-gap matrix $\Delta$ according to the token sequence order. Meanwhile, the temporal positions $t \in \mathbb{R}^{L}$ are defined only along the temporal axis. Each time value is therefore repeated across all variables and rearranged into a one-dimensional sequence aligned with the time-variable cell order to construct the event-aligned timestamp $t^e$:
\begin{equation}
t^e \in \mathbb{R}^{N}, \qquad
t^e = \mathrm{reshape}(t \otimes \mathbf{1}_{V}),
\label{eq:3}
\end{equation}
where $\mathbf{1}_{V}$ is an all-one vector of length $V$, and $\otimes$ denotes repetition of the temporal axis along the variable axis. In addition, a variable index sequence $v^e \in \{0,\dots,V-1\}^{N}$ is defined to indicate the variable associated with each token unit. This preserves variable-specific identity for each element in the flattened sequence and is later used to model variable-dependent freshness decay.

\textbf{Normalization.} Since the temporal position within the observation window and the elapsed time since the previous observation have different meanings and ranges, time-related inputs are normalized for stable integration within a single token representation. Let $T_{\max}$ denote the maximum length of the observation window in minutes. The event-aligned time is linearly transformed from the absolute range $[0, T_{\max}]$ to $[-1,1]$:
\begin{equation}
\tau_i = \mathrm{clip}\!\left(\frac{2 t_i^e}{T_{\max}} - 1,\,-1,\,1\right),
\label{eq:4}
\end{equation}
where $t_i^e$ is the absolute measurement time of the $i$-th token unit, and $\tau_i$ is the normalized temporal position. Here, $\mathrm{clip}(\cdot)$ restricts a value to the specified range. Next, the raw time gap $\Delta_i^e$ may exhibit a distribution skewed toward large values in clinical data. Log scaling is therefore applied to mitigate this heavy-tailed property, after which the result is normalized to the range $[0,1]$:
\begin{equation}
\bar{\Delta}_i
=
\mathrm{clip}\!\left(
\frac{\log(1+\Delta_i^e)}{\log(1+T_{\max})},
\,0,\,1
\right),
\label{eq:5}
\end{equation}
where $\Delta_i^e$ is the raw time gap of the $i$-th token unit, and $\bar{\Delta}_i$ is the normalized staleness magnitude in $[0,1]$. The normalized value is then remapped to $[-1,1]$ so that it can be encoded together with $\tau_i$ on a consistent scale:
\begin{equation}
\delta_i = 2\bar{\Delta}_i - 1,
\label{eq:6}
\end{equation}
where $\delta_i$ is the normalized gap representation used for staleness embedding. By using $\tau_i$ and $\delta_i$ separately, the model can distinguish the absolute temporal position of each token unit from the recency of its information.

For entries with $m_i^e = 0$, the value input is forward-filled from the most recent observation when available; otherwise, it is set to 0 after variable-wise normalization. These inputs are treated as placeholders rather than observed measurements, and their contribution is later modulated by the observation mask and elapsed-time signals through the Reliability Gate.

\textbf{Event Token Embedding.} Each token unit is embedded by combining the normalized temporal representation $\tau_i$, value input $x_i^e$, variable index $v_i^e$, and staleness representation $\delta_i$. The continuous inputs $\tau_i$, $x_i^e$, and $\delta_i$ are transformed through continuous value embeddings, while the variable identity $v_i^e$ is represented by a learnable embedding corresponding to each variable index:
\begin{equation}
\small
\begin{aligned}
e_i^{(t)} &= \mathrm{CVE}_t(\tau_i), \quad
e_i^{(x)} = \mathrm{CVE}_x(x_i^e), \\
e_i^{(v)} &= \mathrm{Emb}(v_i^e), \quad
e_i^{(\Delta)} = \mathrm{CVE}_{\Delta}(\delta_i),
\end{aligned}
\label{eq:7}
\end{equation}
\normalsize
where $e_i^{(t)}$, $e_i^{(x)}$, $e_i^{(v)}$, and $e_i^{(\Delta)} \in \mathbb{R}^{D}$ are independent $D$-dimensional embedding vectors corresponding to the temporal position, value input, variable identity, and staleness signal, respectively. Here, $\mathrm{CVE}_t(\cdot)$, $\mathrm{CVE}_x(\cdot)$, and $\mathrm{CVE}_{\Delta}(\cdot)$ denote embedding functions that map continuous scalar inputs to $D$-dimensional vectors, and $\mathrm{Emb}(\cdot)$ denotes a learnable embedding function for variable indices. The final event embedding is computed as follows:
\begin{equation}
z_i = e_i^{(t)} + e_i^{(x)} + e_i^{(v)} + e_i^{(\Delta)},
\label{eq:8}
\end{equation}
where $z_i \in \mathbb{R}^{D}$ is the final representation of the $i$-th token unit. The full event token embedding sequence is then denoted by $Z$:
\begin{equation}
Z = [z_1,\dots,z_N]^{\top} \in \mathbb{R}^{N \times D},
\label{eq:9}
\end{equation}
where $N$ is the total number of token units within the observation window and $D$ is the embedding dimension. $Z$ serves as the base representation for subsequent stages to preserve time-variable information, incorporate observation reliability, and construct summary tokens across multiple temporal resolutions.

\subsection{Reliability Gate}

The Reliability Gate converts missingness and elapsed time into a continuous reliability weight used during aggregation. Its primary role is to distinguish observed entries from imputed placeholders and to downweight uncertain missing information, rather than to encode all recency effects within the gate itself. To this end, a positive decay rate $\lambda_c$ is learned for each variable channel $c \in \{0,\dots,V-1\}$:
\begin{equation}
\lambda_c = \mathrm{softplus}(w_c) + \lambda_{\min},
\label{eq:10}
\end{equation}
where $w_c$ is a learnable scalar parameter for channel $c$, and $\lambda_c > 0$ is the decay rate of that channel. $\mathrm{softplus}(\cdot)$ enforces positivity, and $\lambda_{\min} > 0$ prevents excessively small decay. For each token unit $i$, the reliability weight is computed using the decay rate $\lambda_{v_i^e}$ corresponding to its variable index $v_i^e$:
\begin{equation}
\mathrm{rel}_{w,i}
=
m_i^e + (1 - m_i^e)\exp\!\left(-\lambda_{v_i^e}\Delta_i^e\right),
\label{eq:11}
\end{equation}
where $\mathrm{rel}_{w,i} \in \mathbb{R}$ is the reliability weight of the $i$-th token unit. If $m_i^e = 1$, the corresponding time-variable entry is observed, and thus $\mathrm{rel}_{w,i}=1$. If $m_i^e = 0$, the entry is missing, and the reliability decreases exponentially as $\Delta_i^e$ increases. Thus, $\lambda_{v_i^e}$ controls the variable-specific decay rate of reliability, while $\Delta_i^e$ reflects the staleness of missing information. Collecting all token-wise weights yields the output of the Reliability Gate:
\begin{equation}
\mathrm{rel}_w = [\mathrm{rel}_{w,1},\dots,\mathrm{rel}_{w,N}]^{\top} \in \mathbb{R}^{N},
\label{eq:12}
\end{equation}
where $\mathrm{rel}_w$ is the reliability vector over the full flattened sequence. It is used in the Multi-scale Aggregation stage to modulate the contribution of each token unit when time-variable entries are aggregated into interval-level summaries. This design places greater emphasis on directly observed data while suppressing uncertain missing information. For observed entries, recency is modeled separately through the $\Delta t$ embedding and the bucket-level mean staleness used in multi-scale aggregation. Accordingly, the gate and staleness-related features play complementary roles rather than encoding the same notion twice.

\subsection{Multi-scale Aggregation}

Multi-scale Aggregation constructs compressed summary-token sequences at multiple temporal resolutions while preserving time-variable information from the flattened input. It consists of five submodules: Bucketing, Weighted Pooling, Stats Augment, Concat Scales, and Chronological Weaving. Let $S=\{s_1,s_2,\dots,s_M\}$ denote the set of temporal resolutions, where each $s \in S$ is a temporal scale in minutes. The specific scale values are given in the experimental setup.

\textbf{Bucketing.} Bucketing assigns each token unit to a time interval at a given temporal scale. To compute the bucket index of the $i$-th token unit at scale $s$, the normalized time $\tau_i$ is first restored to its position in minutes:
\begin{equation}
u_i = \frac{\tau_i + 1}{2}T_{\max},
\label{eq:13}
\end{equation}
where $u_i$ is the restored temporal position in minutes. The bucket index of token unit $i$ at scale $s$ is then defined as
\begin{equation}
b_i^{(s)} = \left\lfloor \frac{u_i}{s} \right\rfloor,
\label{eq:14}
\end{equation}
where $b_i^{(s)}$ is the bucket index and $\lfloor \cdot \rfloor$ denotes the floor operation. Thus, bucket assignment follows a left-closed, right-open convention, so that each token unit is assigned to exactly one interval according to its restored time position. The set of indices assigned to bucket $k$ at scale $s$ is
\begin{equation}
B_k^{(s)} = \{\, i \mid b_i^{(s)} = k \,\},
\label{eq:15}
\end{equation}
where $B_k^{(s)}$ is the index set of token units in bucket $k$ at scale $s$. In addition, the center time of each bucket is defined as
\begin{equation}
\theta_k^{(s)} = \left(k + \frac{1}{2}\right)s,
\label{eq:16}
\end{equation}
where $\theta_k^{(s)}$ is the center time of bucket $k$ at scale $s$ and is later used for temporal ordering across scales.

\textbf{Weighted Pooling.} Weighted Pooling aggregates token units within each bucket in a reliability-aware manner to form a single vector representation for the corresponding interval:
\begin{equation}
\mu_k^{(s)}
=
\frac{\sum_{i \in B_k^{(s)}} \mathrm{rel}_{w,i} z_i}
{\sum_{i \in B_k^{(s)}} \mathrm{rel}_{w,i} + \varepsilon},
\label{eq:17}
\end{equation}
where $\mu_k^{(s)} \in \mathbb{R}^{D}$ is the pooled representation of bucket $k$ at scale $s$, $z_i$ is the $i$-th token embedding, $\mathrm{rel}_{w,i}$ is its reliability weight, and $\varepsilon > 0$ is a small constant to prevent division by zero. If $B_k^{(s)}$ is empty, no token is generated for that bucket and it is excluded in later concatenation.

\textbf{Stats Augment.} To enrich mean-pooled bucket representations, this module incorporates a dispersion signal and bucket-level statistics that summarize observation density and freshness. The dispersion signal $\nu_k^{(s)}$ is computed from the reliability-weighted second moment:
\begin{equation}
\nu_k^{(s)}
=
\left(
\frac{\sum_{i \in B_k^{(s)}} \mathrm{rel}_{w,i}(z_i \odot z_i)}
{\sum_{i \in B_k^{(s)}} \mathrm{rel}_{w,i} + \varepsilon}
-
\mu_k^{(s)} \odot \mu_k^{(s)}
\right)_{+},
\label{eq:18}
\end{equation}
where $\nu_k^{(s)} \in \mathbb{R}^{D}$ reflects variance-like feature behavior within the bucket, $\odot$ denotes element-wise product, and $(\cdot)_{+}$ denotes element-wise nonnegative clipping. The dispersion information is then transformed in a scale-specific manner and added to the bucket representation:
\begin{equation}
\mu_k^{(s)} \leftarrow \mu_k^{(s)} + \phi_s\!\left(\nu_k^{(s)}\right),
\label{eq:19}
\end{equation}
where $\phi_s(\cdot)$ is a scale-specific transformation implemented with layer normalization and a learnable projection.

Bucket-level measurement patterns are also incorporated using observation count, coverage, effective count, and mean staleness:
\begin{equation}
n_{k,\mathrm{obs}}^{(s)} = \sum_{i \in B_k^{(s)}} m_i^e,
\qquad
n_{k,\mathrm{all}}^{(s)} = \sum_{i \in B_k^{(s)}} 1,
\label{eq:20}
\end{equation}
where $n_{k,\mathrm{obs}}^{(s)}$ and $n_{k,\mathrm{all}}^{(s)}$ denote the numbers of observed and total entries, respectively. Since empty buckets are removed in advance, $n_{k,\mathrm{all}}^{(s)} \geq 1$. Coverage is defined as
\begin{equation}
\rho_k^{(s)} = \frac{n_{k,\mathrm{obs}}^{(s)}}{n_{k,\mathrm{all}}^{(s)}},
\label{eq:21}
\end{equation}
where $\rho_k^{(s)} \in [0,1]$ is the proportion of observed entries in the bucket. The reliability-weighted effective count $w_k^{(s)}$ and mean staleness $\bar{\Delta}_k^{(s)}$ are defined as
\begin{equation}
w_k^{(s)} = \sum_{i \in B_k^{(s)}} \mathrm{rel}_{w,i},
\label{eq:22}
\end{equation}
\begin{equation}
\bar{\Delta}_k^{(s)}
=
\frac{\sum_{i \in B_k^{(s)}} \mathrm{rel}_{w,i}\bar{\Delta}_i}
{\sum_{i \in B_k^{(s)}} \mathrm{rel}_{w,i} + \varepsilon},
\label{eq:23}
\end{equation}
and these statistics are concatenated as
\begin{equation}
g_k^{(s)}
=
\left[
\log\!\bigl(1+n_{k,\mathrm{obs}}^{(s)}\bigr),
\;
\rho_k^{(s)},
\;
\log\!\bigl(1+w_k^{(s)}\bigr),
\;
\bar{\Delta}_k^{(s)}
\right],
\label{eq:24}
\end{equation}
where $g_k^{(s)} \in \mathbb{R}^{4}$ is the bucket-level statistics vector. The statistics embedding produced by an MLP is added to the bucket token. Each bucket also receives a bucket-center-time embedding and a scale embedding to preserve temporal position and resolution information.

\textbf{Concat Scales and Chronological Weaving.} This module integrates the summary tokens generated at each scale and reorders them along the temporal axis so that the backbone can process multi-resolution information in a more coherent temporal context. Let $Z^{(s)}$ denote the valid bucket-token sequence generated at scale $s \in S=\{s_1,\dots,s_M\}$. The full multi-scale token sequence, together with its corresponding mask and bucket-center-time sequence, is defined as
\begin{equation}
Z = [Z^{(s_1)}; Z^{(s_2)}; \dots; Z^{(s_M)}] \in \mathbb{R}^{N_{\mathrm{tok}} \times D},
\label{eq:25}
\end{equation}
\begin{equation}
M = [M^{(s_1)}; M^{(s_2)}; \dots; M^{(s_M)}] \in \{0,1\}^{N_{\mathrm{tok}}},
\label{eq:26}
\end{equation}
\begin{equation}
T = [T^{(s_1)}; T^{(s_2)}; \dots; T^{(s_M)}] \in \mathbb{R}^{N_{\mathrm{tok}}},
\label{eq:27}
\end{equation}
where $Z^{(s)}$ is the sequence of valid bucket tokens at scale $s$, and $M^{(s)}$ and $T^{(s)}$ are the corresponding valid mask and ordered sequence of bucket center times $\theta_k^{(s)}$ defined in \eqref{eq:16}. Although empty buckets are removed in advance, the valid-token mask $M$ is retained for notational consistency with later token selection and sequence processing. Here, $D$ is the token embedding dimension, and $N_{\mathrm{tok}}$ is the total number of valid tokens across all scales.

Simple concatenation arranges tokens in scale-wise blocks, so summaries describing the same or nearby periods at different resolutions may be far apart in the sequence. To address this, Chronological Weaving reorders all tokens according to $T$, so that summaries centered around similar time points become adjacent regardless of resolution. Because each bucket token is constructed only from entries within its own interval and weaving is applied only after summary construction, this step does not mix raw observations across intervals. As a result, the backbone can process nearby cross-scale summaries in a coherent time-ordered context without additional cross-interval aggregation.

\subsection{Sequence Modeling}

The Sequence Modeling stage takes as input the multi-scale token sequence $Z$, the corresponding mask $M$, and the temporal position sequence $T$. Its goal is to compress the multi-scale token sequence under a fixed budget and encode it with a Mamba backbone for final prediction.

\textbf{Budgeted Token Router.} Although multi-scale aggregation summarizes information across multiple temporal resolutions, the resulting token sequence can still be long depending on the number of scales and buckets. To address this, a budgeted token router is applied before sequence encoding. This module reduces computation while preserving predictive information by assigning routing scores to tokens and retaining a compact subset under a fixed token budget. The importance score $s_j$ of each token $Z_j$ is computed as follows:
\begin{equation}
s_j = w_r^{\top} Z_j + b_r,
\label{eq:28}
\end{equation}
where $w_r \in \mathbb{R}^{D}$ and $b_r \in \mathbb{R}$ are learnable router parameters. During training, the routing scores are converted into differentiable soft routing weights, which are used to modulate token representations so that the router can be optimized jointly with the downstream prediction objective. During inference, hard top-$k$ selection is applied based on the routing scores to construct a compact sequence with a fixed token budget. Here, $k$ is a hyperparameter. In this way, soft routing is used to maintain differentiability during training, whereas hard top-$k$ routing is used to enforce fixed-budget sequence compression at inference time.

The selected tokens are then reordered chronologically to form $(Z', M', T')$. Here, $Z' \in \mathbb{R}^{N' \times D}$ is the selected token sequence, $M' \in \{0,1\}^{N'}$ is the corresponding mask, and $T' \in \mathbb{R}^{N'}$ contains the temporal positions of the selected tokens. In addition, $N' \leq k$ denotes the number of valid tokens after routing. Since routing is based on token importance, the selected order may not follow time. Chronological reordering restores a time-ordered sequence before Mamba encoding.

\textbf{Mamba Backbone and Prediction.} The reordered token sequence $Z'$ is fed into the Mamba backbone. Mamba \cite{ref13} is used as the sequence encoder to efficiently capture temporal dependencies among the selected summary tokens. The backbone output sequence is written as follows:
\begin{equation}
H = \mathrm{Mamba}(Z') \in \mathbb{R}^{N' \times D},
\label{eq:29}
\end{equation}
where $H$ is the backbone output sequence, and $h_j \in \mathbb{R}^{D}$ is the $j$-th output token representation. The final prediction uses the representation of the last valid token:
\begin{equation}
h_{\mathrm{pool}} = h_{N'},
\label{eq:30}
\end{equation}
where $h_{\mathrm{pool}} \in \mathbb{R}^{D}$ is the sample-level pooled representation. After chronological reordering, the last token corresponds to the latest retained summary in the reordered sequence. This design preserves the most recent summarized patient state after temporal reordering while keeping the prediction head simple. The logit and prediction are then computed through a linear output head:
\begin{equation}
\mathrm{logit} = W_o h_{\mathrm{pool}} + b_o,
\qquad
\hat{y} = \sigma(\mathrm{logit}),
\label{eq:31}
\end{equation}
where $W_o$ and $b_o$ are output-head parameters, $\mathrm{logit} \in \mathbb{R}$ is the pre-sigmoid score for binary classification, and $\hat{y} \in (0,1)$ is the final predicted probability.

\section{Experiments}

This section evaluates ReTAMamba on three clinical time-series benchmarks. After describing the experimental setup, it presents overall performance comparisons, component-wise ablations, and efficiency analysis. It then examines differences in observation patterns between survivors and non-survivors to interpret model behavior, and finally quantifies the effect of the multi-resolution temporal design.

\subsection{Experimental Setup}

\textbf{Datasets.} Experiments were conducted on three ICU clinical time-series benchmarks with irregular sampling and missing values: MIMIC-IV \cite{ref20}, eICU \cite{ref21}, and PhysioNet 2012 \cite{ref22}. The prediction task was unified as in-hospital mortality prediction across all datasets. For each patient, only the first 48 hours after ICU admission were used as input to reflect an early risk assessment setting. The input consisted of 17 clinical variables commonly used in prior clinical time-series studies, and only variables available across all three datasets were retained to ensure a consistent evaluation setting \cite{ref3}. Cases with missing outcome labels, invalid ICU stay information, or insufficient observations within the 48-hour window were excluded during preprocessing. These datasets differ in cohort scale, missingness patterns, and observation density, providing complementary testbeds for evaluating robustness under a unified early-risk prediction setting across heterogeneous ICU environments.

\textbf{Baselines.} To evaluate the proposed design from multiple perspectives, the baselines were organized into three groups. First, we included general-purpose predictive backbones, namely XGBoost \cite{ref23}, LSTM \cite{ref24}, Transformer \cite{ref12}, and Mamba \cite{ref13}. Second, we considered clinical time-series models designed to handle irregularity and missingness, including GRU-D \cite{ref5}, IP-Nets \cite{ref7}, mTAN \cite{ref11}, SeFT \cite{ref9}, and Raindrop \cite{ref8}. Third, we included Warpformer \cite{ref10} as the most relevant comparator because it explicitly combines temporal aggregation with multi-scale representation learning for irregular clinical time series.

\textbf{Evaluation Metrics.} AUROC and AUPRC were used to evaluate binary classification performance. Because in-hospital mortality prediction is highly imbalanced, AUPRC was treated as the primary metric.

\begin{table}[!htbp]
\centering
\caption{Temporal input types used by different models.}
\label{tab:input_types}
\footnotesize
\begin{tabular}{lccc}
\toprule
Model & Mask & Time gap & Timestamp \\
\midrule
XGBoost      & \checkmark & --         & --         \\
LSTM         & \checkmark & \checkmark & --         \\
Transformer  & \checkmark & \checkmark & --         \\
Mamba        & \checkmark & \checkmark & --         \\
GRU-D        & \checkmark & \checkmark & --         \\
IP-Nets      & \checkmark & \checkmark & --         \\
mTAN         & \checkmark & \checkmark & \checkmark \\
SeFT         & \checkmark & \checkmark & \checkmark \\
Raindrop     & \checkmark & \checkmark & \checkmark \\
Warpformer   & \checkmark & \checkmark & \checkmark \\
ReTAMamba    & \checkmark & \checkmark & \checkmark \\
\bottomrule
\end{tabular}
\end{table}

\begin{table*}[t]
\centering
\footnotesize
\caption{Main results on three clinical time-series benchmarks.}
\label{tab:main_results}
\setlength{\tabcolsep}{5pt}
\renewcommand{\arraystretch}{1.08}

\begin{tabularx}{\textwidth}{l *{6}{>{\centering\arraybackslash}Y}}
\toprule
\multirow{2}{*}{Models} & \multicolumn{2}{c}{MIMIC-IV} & \multicolumn{2}{c}{eICU} & \multicolumn{2}{c}{PhysioNet 2012} \\
\cmidrule(lr){2-3} \cmidrule(lr){4-5} \cmidrule(lr){6-7}
& AUROC & AUPRC & AUROC & AUPRC & AUROC & AUPRC \\
\midrule
XGBoost     & $0.815 \pm 0.001$ & $0.481 \pm 0.002$ & $0.764 \pm 0.001$ & $0.343 \pm 0.001$ & $0.804 \pm 0.001$ & $0.443 \pm 0.001$ \\
LSTM        & $0.850 \pm 0.002$ & $0.515 \pm 0.001$ & $0.781 \pm 0.003$ & $0.350 \pm 0.004$ & $0.826 \pm 0.005$ & $0.458 \pm 0.017$ \\
Transformer & $0.852 \pm 0.003$ & $0.523 \pm 0.005$ & $0.784 \pm 0.002$ & $0.361 \pm 0.001$ & $0.820 \pm 0.008$ & $0.468 \pm 0.012$ \\
Mamba       & $0.840 \pm 0.002$ & $0.490 \pm 0.008$ & $0.774 \pm 0.004$ & $0.351 \pm 0.006$ & $0.814 \pm 0.007$ & $0.468 \pm 0.014$ \\
GRU-D       & $0.854 \pm 0.002$ & $0.517 \pm 0.005$ & $0.783 \pm 0.004$ & $0.358 \pm 0.007$ & $0.831 \pm 0.003$ & $0.487 \pm 0.011$ \\
IP-Nets     & $0.856 \pm 0.002$ & $0.529 \pm 0.004$ & $0.787 \pm 0.004$ & $0.362 \pm 0.003$ & $0.825 \pm 0.004$ & $0.463 \pm 0.007$ \\
mTAN        & $0.848 \pm 0.009$ & $0.511 \pm 0.008$ & $0.778 \pm 0.005$ & $0.354 \pm 0.005$ & $0.829 \pm 0.006$ & $0.472 \pm 0.010$ \\
SeFT        & $0.853 \pm 0.001$ & $0.514 \pm 0.005$ & $0.783 \pm 0.005$ & $0.359 \pm 0.009$ & $0.833 \pm 0.005$ & $0.501 \pm 0.013$ \\
Raindrop    & $0.841 \pm 0.003$ & $0.489 \pm 0.010$ & $0.775 \pm 0.005$ & $0.352 \pm 0.007$ & $0.821 \pm 0.003$ & $0.476 \pm 0.009$ \\
Warpformer  & $0.855 \pm 0.004$ & $0.528 \pm 0.014$ & $0.784 \pm 0.002$ & $0.372 \pm 0.004$ & $0.831 \pm 0.007$ & $0.485 \pm 0.020$ \\
\textbf{ReTAMamba}   & $\mathbf{0.861 \pm 0.001}$ & $\mathbf{0.548 \pm 0.005}$ & $\mathbf{0.793 \pm 0.004}$ & $\mathbf{0.384 \pm 0.003}$ & $\mathbf{0.847 \pm 0.002}$ & $\mathbf{0.520 \pm 0.013}$ \\
\bottomrule
\end{tabularx}
\vspace{2pt}
\noindent{\raggedright \textbf{Note.} Best results are shown in bold. All results are reported as mean $\pm$ standard deviation over five random seeds.\par}
\end{table*}

\begin{table*}[t]
\centering
\footnotesize
\caption{Ablation results of key components in ReTAMamba on three clinical time-series benchmarks.}
\label{tab:ablation_results}
\setlength{\tabcolsep}{5pt}
\renewcommand{\arraystretch}{1.08}
\begin{tabularx}{\textwidth}{l *{6}{>{\centering\arraybackslash}Y}}
\toprule
\multirow{2}{*}{Variant} & \multicolumn{2}{c}{MIMIC-IV} & \multicolumn{2}{c}{eICU} & \multicolumn{2}{c}{PhysioNet 2012} \\
\cmidrule(lr){2-3} \cmidrule(lr){4-5} \cmidrule(lr){6-7}
& AUROC & AUPRC & AUROC & AUPRC & AUROC & AUPRC \\
\midrule
\textbf{ReTAMamba (Full)} & $\mathbf{0.861 \pm 0.001}$ & $\mathbf{0.548 \pm 0.005}$ & $\mathbf{0.793 \pm 0.004}$ & $\mathbf{0.384 \pm 0.003}$ & $\mathbf{0.847 \pm 0.002}$ & $\mathbf{0.520 \pm 0.013}$ \\
w/o Reliability Gate      & $0.850 \pm 0.003$ & $0.516 \pm 0.012$ & $0.784 \pm 0.002$ & $0.367 \pm 0.004$ & $0.823 \pm 0.004$ & $0.487 \pm 0.015$ \\
w/o $\Delta t$ Embedding  & $0.858 \pm 0.002$ & $0.528 \pm 0.007$ & $0.786 \pm 0.003$ & $0.369 \pm 0.003$ & $0.842 \pm 0.005$ & $0.508 \pm 0.014$ \\
w/o Stats Augment         & $0.855 \pm 0.003$ & $0.535 \pm 0.006$ & $0.787 \pm 0.003$ & $0.369 \pm 0.004$ & $0.841 \pm 0.004$ & $0.507 \pm 0.012$ \\
w/o Token Router          & $0.858 \pm 0.002$ & $0.538 \pm 0.006$ & $0.780 \pm 0.002$ & $0.361 \pm 0.004$ & $0.841 \pm 0.003$ & $0.512 \pm 0.010$ \\
w/o Chronological Weaving & $0.847 \pm 0.003$ & $0.520 \pm 0.005$ & $0.775 \pm 0.003$ & $0.352 \pm 0.004$ & $0.826 \pm 0.007$ & $0.501 \pm 0.010$ \\
\bottomrule
\end{tabularx}
\vspace{2pt}
\noindent{\raggedright \textbf{Note.} Each variant removes one key component from ReTAMamba. Best results are shown in bold. All results are reported as mean $\pm$ standard deviation over five random seeds.\par}
\end{table*}

\textbf{Implementation Details.} All models were trained and evaluated under the same task definition, patient-level train/validation/test splits (0.70/0.10/0.20), 48-hour observation window, variable set, and preprocessing pipeline to ensure fair comparison. The temporal input types used by different models are summarized in Table~\ref{tab:input_types}. The shared preprocessing pipeline was applied to the value inputs of all models, while temporal inputs such as masks, time gaps, and timestamps were additionally provided only when required by each architecture. Missing values were forward-filled from the most recent observation when available; unresolved values were then set to 0 after variable-wise z-score normalization using training-set statistics, while missingness and elapsed time were preserved through mask and time-gap features. All neural models were optimized with AdamW using a batch size of 64 and trained for up to 50 epochs with early stopping based on validation performance; the best validation checkpoint was used for test evaluation. Results are reported as averages over five random seeds. Hyperparameters for all models were tuned on the validation set using Optuna under matched search budgets. The final hyperparameter settings of ReTAMamba are summarized in Appendix~A.1 (Table~\ref{tab:final_hyperparameters}).

\vspace{-0.8em}

\subsection{Main Results}

This section quantitatively compares the overall predictive performance of ReTAMamba against existing models. Table~\ref{tab:main_results} summarizes the overall prediction results on three clinical time-series benchmarks. ReTAMamba achieves the best AUROC and AUPRC on all three datasets, namely MIMIC-IV, eICU, and PhysioNet 2012, demonstrating consistently strong discriminative performance across diverse clinical time-series settings.

Particular emphasis is placed on AUPRC, which is more informative than AUROC under severe class imbalance. Under this metric, ReTAMamba improves the average AUPRC of all baseline models by 7.51\%, 7.80\%, and 10.15\% on MIMIC-IV, eICU, and PhysioNet 2012, respectively. Compared with the strongest baseline, Warpformer, it also achieves absolute AUPRC gains of 0.020, 0.012, and 0.035. A paired t-test against the strongest competing model on each dataset showed that the AUPRC improvements were statistically significant on all three datasets ($p<0.05$). These results support the effectiveness of jointly modeling observation reliability, recency, multi-scale temporal context, and budgeted sequence compression for irregular clinical time-series prediction. A supplementary calibration analysis on the eICU dataset is provided in Table~\ref{tab:calibration_results}.

\vspace{-1.4em}

\subsection{Ablation Study}

This section analyzes the contribution of each component in ReTAMamba through ablation experiments that remove one key module at a time. Table~\ref{tab:ablation_results} shows that the full ReTAMamba achieves the best performance on all three datasets.

Removing any single component leads to performance degradation, indicating that each design element contributes to the overall effectiveness of the model. In particular, notable drops are observed in w/o Reliability Gate and w/o Chronological Weaving, highlighting the importance of reliability-aware weighting and chronological ordering of multi-resolution tokens. The degradation is especially clear in AUPRC, which is more sensitive to class imbalance in clinical prediction tasks.

To further validate the role of Chronological Weaving, we additionally compared it with random reordering on eICU. As shown in Table~\ref{tab:weaving_modes}, chronological ordering achieved the best performance, outperforming both no weaving and random reordering. This suggests that the benefit of the module comes not merely from reordering tokens, but from arranging multi-resolution summaries in a coherent time-ordered manner.

Performance also decreases in w/o Stats Augment across all three datasets, suggesting that bucket-level statistics, such as within-bucket variance, observation frequency, coverage, and mean staleness, provide useful summary information for irregular clinical time series. In addition, w/o $\Delta t$ Embedding also shows consistent performance reductions, indicating the importance of encoding recency in the event representation. The performance drop in w/o Token Router further suggests that budgeted token routing helps preserve predictive information while controlling sequence length before Mamba encoding. Supplementary statistics on pre-routing token counts and token budget sensitivity are provided in Table~\ref{tab:pre_routing_tokens} and Table~\ref{tab:k_budget_sensitivity}, respectively.

\begin{table}[h]
\centering
\caption{Effect of Chronological Weaving on eICU.}
\label{tab:weaving_modes}
\footnotesize
\begin{tabular}{lcc}
\toprule
Weaving mode & AUROC & AUPRC \\
\midrule
Chronological      & $0.793 \pm 0.004$  & $0.384 \pm 0.003$ \\
None (w/o weaving) & $0.775 \pm 0.003$  & $0.352 \pm 0.004$ \\
Random             & $0.784 \pm 0.002$  & $0.364 \pm 0.003$ \\
\bottomrule
\end{tabular}
\end{table}

\vspace{-0.8em}

\subsection{Efficiency Analysis}

This section evaluates the computational efficiency of ReTAMamba in terms of parameter count, training peak memory, training step time, and inference latency, in comparison with existing models on eICU.

\begin{figure}[h!]
\centering
\begin{minipage}[t]{0.48\linewidth}
    \centering
    \includegraphics[width=\linewidth]{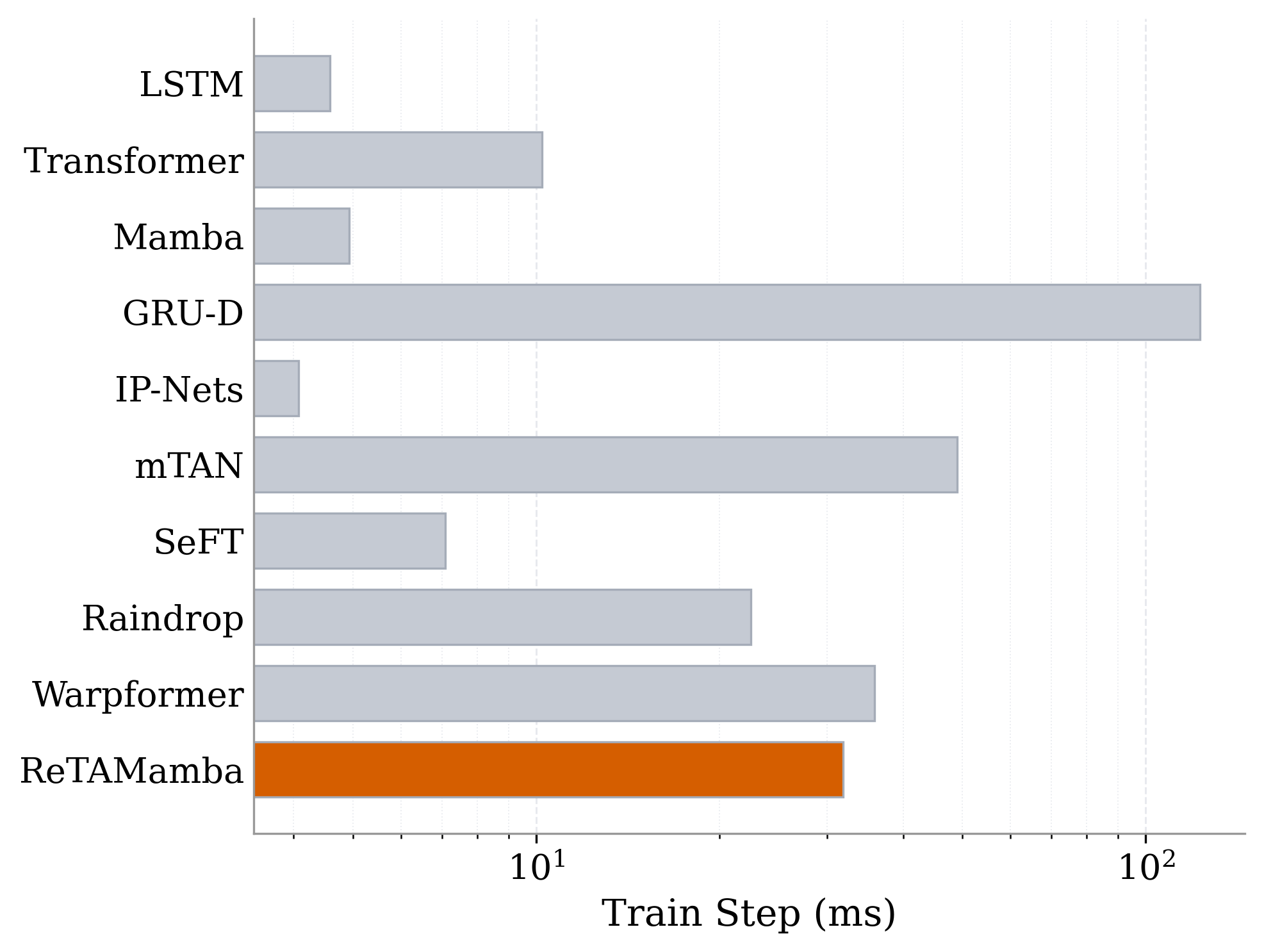}
    
    (a) Training step time
\end{minipage}
\hfill
\begin{minipage}[t]{0.48\linewidth}
    \centering
    \includegraphics[width=\linewidth]{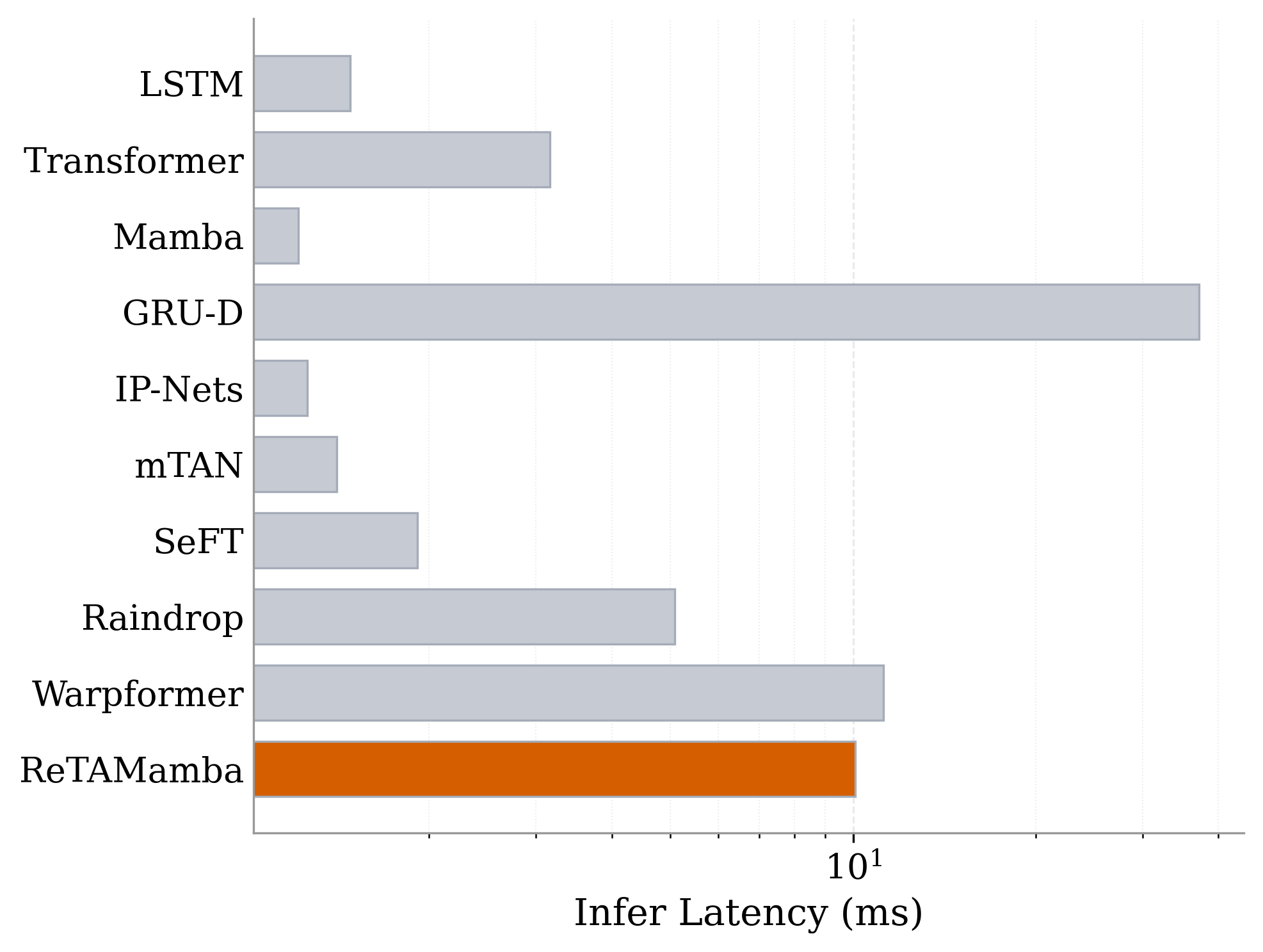}
    
    (b) Inference latency
\end{minipage}
\caption{Efficiency comparison on the eICU dataset.}
\label{fig:efficiency}
\end{figure}

Table~\ref{tab:efficiency} and Figure~\ref{fig:efficiency} summarize the efficiency results. Training peak memory was measured as the maximum allocated GPU memory during training. Training step time included forward pass, backward pass, and optimizer update, and inference latency was evaluated on test samples with batch size 1 under the same implementation and execution setting. All results were measured on a single NVIDIA RTX PRO 5000 Blackwell 48 GB GPU using CUDA with bf16 automatic mixed precision. ReTAMamba maintains moderate computational cost with 157K parameters and 163 MB of training peak memory, while Warpformer shows the highest resource consumption with 240K parameters and 844 MB. Raindrop also requires more memory than ReTAMamba.

\begin{table}[h!]
\centering
\caption{Efficiency comparison of different models.}
\label{tab:efficiency}
\setlength{\tabcolsep}{6pt}
\renewcommand{\arraystretch}{1.08}
\footnotesize
\begin{tabular}{lcc}
\hline
Model & Params (K) & Train Peak Mem (MB) \\
\hline
LSTM       & 63  & 54  \\
Transformer& 70  & 38  \\
Mamba      & 36  & 32  \\
GRU-D      & 20  & 31  \\
IP-Nets    & 23  & 49  \\
mTAN       & 18  & 55  \\
SeFT       & 20  & 99  \\
Raindrop   & 113 & 199 \\
Warpformer & 240 & 844 \\
ReTAMamba  & 157 & 163 \\
\hline
\end{tabular}
\end{table}

As shown in Figure~\ref{fig:efficiency}, ReTAMamba maintains lower training step time and inference latency than GRU-D, mTAN, Raindrop, and Warpformer, although it remains slower than simpler models such as LSTM, Transformer, Mamba, and IP-Nets. These results should be interpreted as controlled wall-clock measurements, since actual efficiency depends not only on parameter count but also on model-specific computation patterns for irregular-sequence processing. Combined with the accuracy results in Table~\ref{tab:main_results}, these findings suggest that ReTAMamba achieves the best predictive performance while maintaining competitive computational efficiency.

\subsection{Temporal Patterns and Model Behavior}

\subsubsection{Cohort Differences}

This section examines survivor-non-survivor differences in observation patterns during the first 48 hours after ICU admission on eICU. To highlight characteristics more pronounced in deceased patients, each metric is visualized as non-survivors minus survivors.

\begin{figure}[h]
\centering
\begin{minipage}[t]{0.48\columnwidth}
    \centering
    \includegraphics[width=\linewidth]{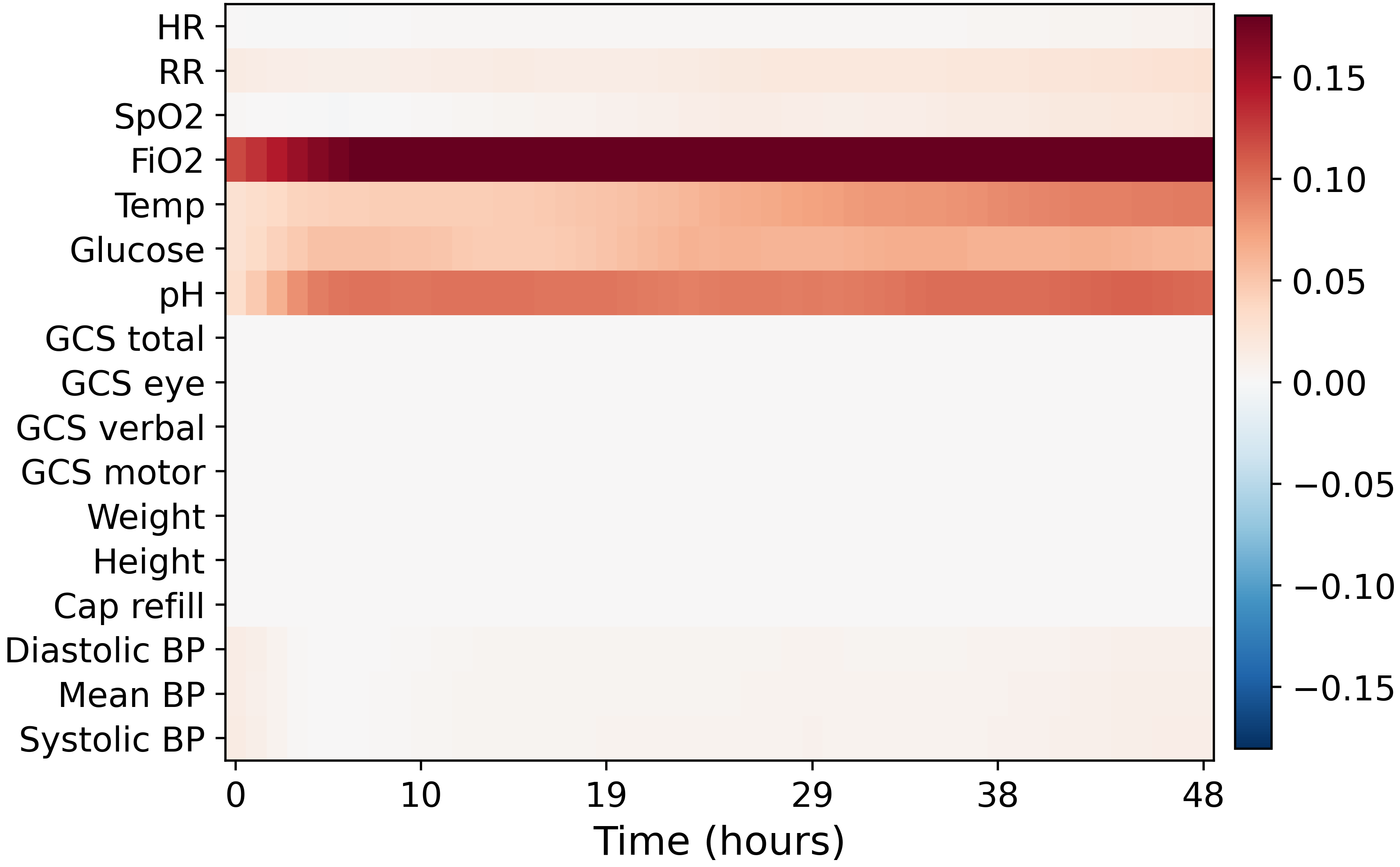}
    
    (a) $\Delta$ Coverage
\end{minipage}
\hfill
\begin{minipage}[t]{0.48\columnwidth}
    \centering
    \includegraphics[width=\linewidth]{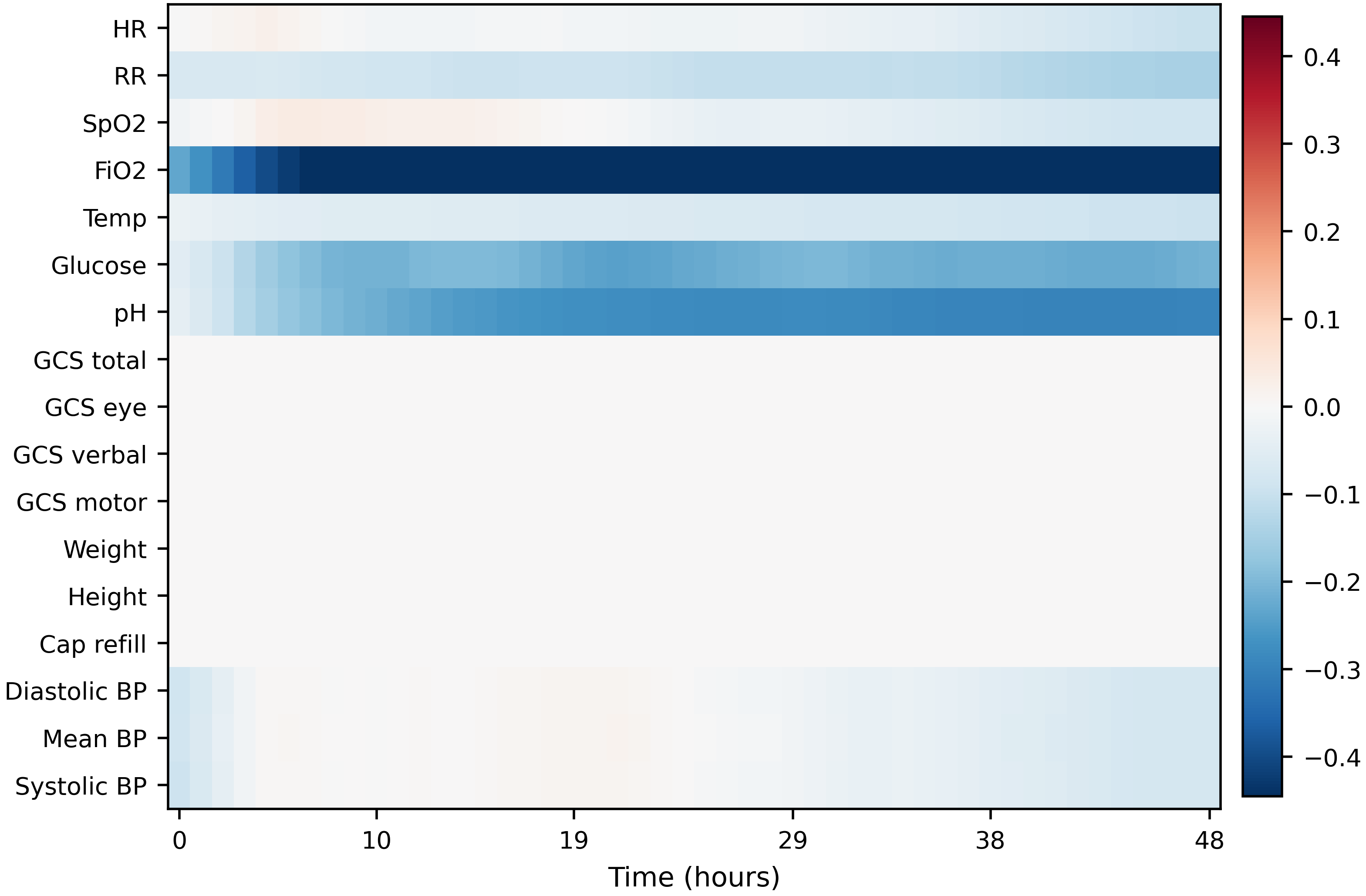}
    
    (b) $\Delta$ Staleness
\end{minipage}
\caption{Temporal cohort differences in the eICU dataset.}
\label{fig:cohort_diff}
\end{figure}

Figure~\ref{fig:cohort_diff} shows the temporal mean differences in coverage and staleness. Positive values indicate larger values in non-survivors, whereas negative values indicate larger values in survivors. Higher $\Delta$coverage indicates more frequent observation in non-survivors, whereas lower $\Delta$staleness indicates more recent observation.

Non-survivors generally exhibited higher coverage together with lower staleness for several variables, particularly FiO$_2$, pH, body temperature, and blood pressure. This suggests that these variables were monitored more frequently and more recently in non-survivors, consistent with intensified clinical attention to respiratory status, acid-base balance, and hemodynamic stability in critically ill patients \cite{ref16,ref25}.

Overall, these results suggest that predictive signals reside not only in measured values but also in the observation process itself. In other words, patient risk may be reflected not only in the blood pressure or pH value itself, but also in how often and how recently clinicians chose to measure it. The temporal differences in coverage and staleness therefore directly motivate the need to model missingness and information freshness jointly in irregular clinical time series \cite{ref26}.

\begin{table}[h]
\centering
\caption{Group-wise summary of learned decay patterns on eICU.}
\label{tab:decay_group_summary}
\small
\begin{tabular}{lccc}
\toprule
Group & Mean decay & Mean coverage & Mean gap (h) \\
\midrule
Vital & 0.2044 & 0.6341 & 1.21 \\
Lab   & 0.1644 & 0.1136 & 6.25 \\
\bottomrule
\end{tabular}
\end{table}

\begin{table*}[t]
\centering
\footnotesize
\caption{Performance comparison across different multi-scale configurations.}
\label{tab:multiscale_effects}
\begin{tabular}{lcccccc}
\toprule
\multirow{2}{*}{Temporal Scales (min)} 
& \multicolumn{2}{c}{MIMIC-IV} 
& \multicolumn{2}{c}{eICU} 
& \multicolumn{2}{c}{PhysioNet 2012} \\
\cmidrule(lr){2-3} \cmidrule(lr){4-5} \cmidrule(lr){6-7}
& AUROC & AUPRC & AUROC & AUPRC & AUROC & AUPRC \\
\midrule
Base: \{60\} 
& 0.854 $\pm$ 0.003 & 0.530 $\pm$ 0.007 
& 0.783 $\pm$ 0.003 & 0.366 $\pm$ 0.005 
& 0.842 $\pm$ 0.006 & 0.503 $\pm$ 0.016 \\
Medium: \{60, 120\} 
& 0.856 $\pm$ 0.003 & 0.541 $\pm$ 0.004 
& 0.784 $\pm$ 0.004 & 0.371 $\pm$ 0.006 
& 0.842 $\pm$ 0.004 & 0.513 $\pm$ 0.012 \\
\textbf{Long: \{60, 120, 240\}} 
& \textbf{0.861 $\pm$ 0.001} & \textbf{0.548 $\pm$ 0.005} 
& \textbf{0.793 $\pm$ 0.004} & \textbf{0.384 $\pm$ 0.003} 
& \textbf{0.847 $\pm$ 0.002} & \textbf{0.520 $\pm$ 0.013} \\
Extended: \{60, 120, 240, 480\} 
& 0.857 $\pm$ 0.004 & 0.540 $\pm$ 0.008 
& 0.784 $\pm$ 0.002 & 0.373 $\pm$ 0.002 
& 0.839 $\pm$ 0.005 & 0.509 $\pm$ 0.011 \\
\bottomrule
\end{tabular}
\end{table*}

\subsubsection{Learned Reliability Patterns}

To examine how the Reliability Gate models information freshness, Table~\ref{tab:decay_group_summary} summarizes the learned decay patterns across major feature groups on eICU. Frequently measured bedside signals in the Vital group, such as heart rate, respiratory rate, blood pressure, and temperature, showed larger decay values, whereas laboratory variables in the Lab group, such as pH and glucose, showed smaller decay values and substantially longer observation gaps. This suggests that the gate learned to reduce the reliability of rapidly outdated signals more aggressively, while preserving the usefulness of more slowly updated variables for longer periods.

\subsubsection{Case Analysis}

To illustrate how these reliability patterns appear at the individual-patient level, representative survivor and non-survivor cases were selected from correctly classified test samples whose predicted probabilities were closest to the class-wise median, in order to avoid cherry-picking highly atypical examples.

\begin{figure}[h]
\centering
\begin{minipage}[t]{0.48\linewidth}
    \centering
    \includegraphics[width=\linewidth]{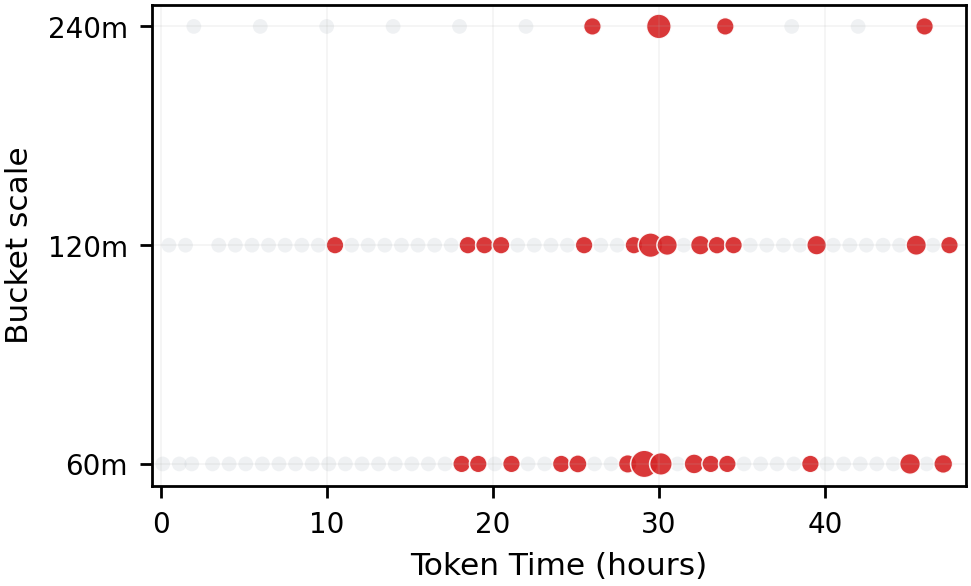}
    
    (a) Survivor Tokens
\end{minipage}
\hfill
\begin{minipage}[t]{0.48\linewidth}
    \centering
    \includegraphics[width=\linewidth]{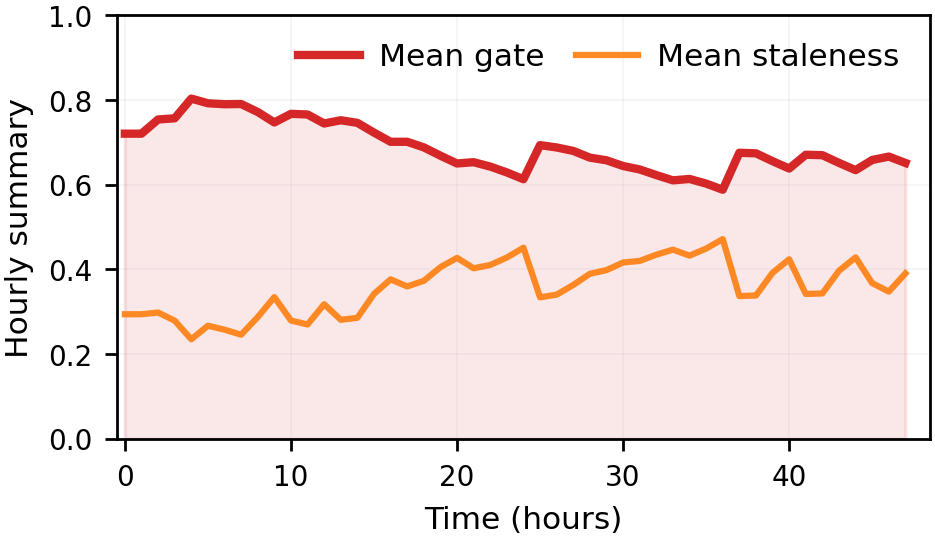}
    
    (b) Survivor Gate/Staleness
\end{minipage}

\vspace{4pt}

\begin{minipage}[t]{0.48\linewidth}
    \centering
    \includegraphics[width=\linewidth]{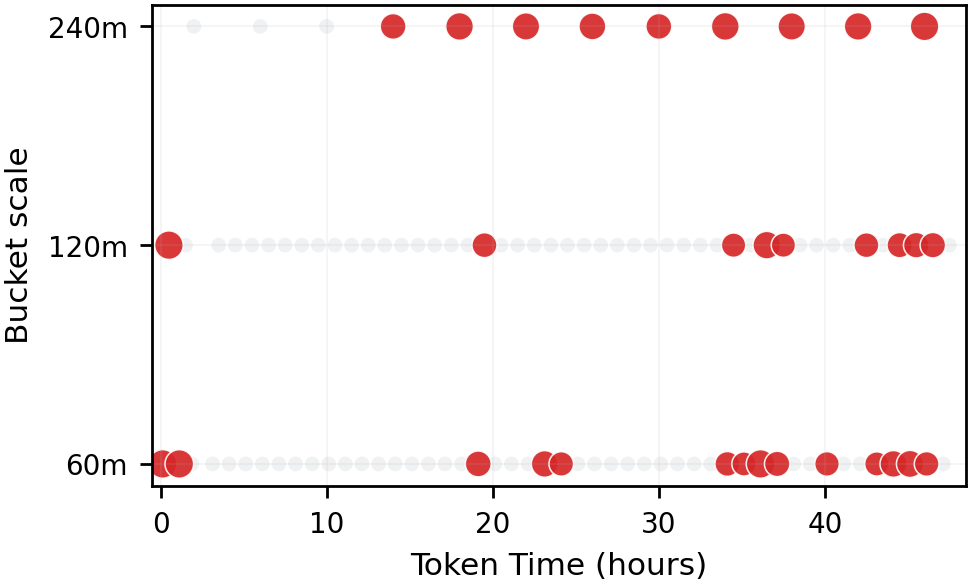}
    
    (c) Non-survivor Tokens
\end{minipage}
\hfill
\begin{minipage}[t]{0.48\linewidth}
    \centering
    \includegraphics[width=\linewidth]{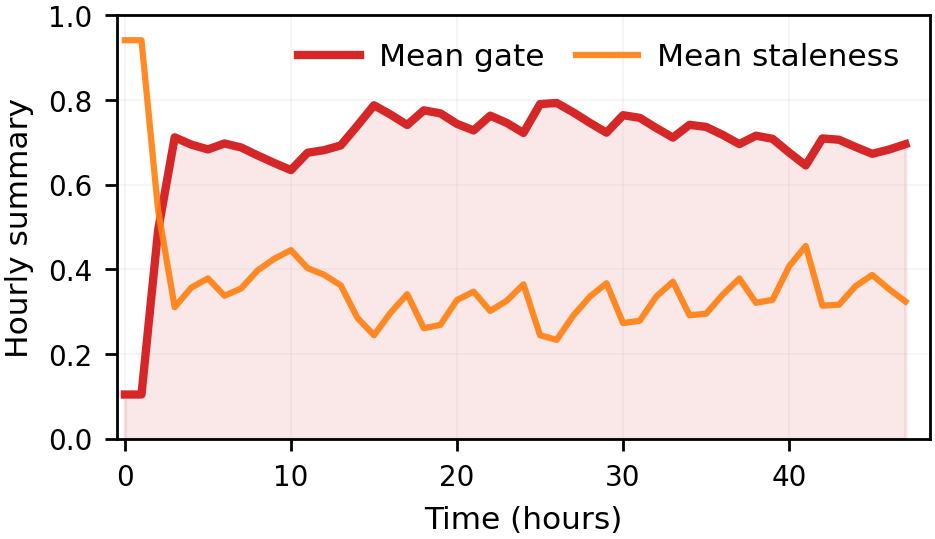}
    
    (d) Non-survivor Gate/Staleness
\end{minipage}
\caption{Case study of model behavior for survivor (a,b) and non-survivor (c,d).}
\label{fig:case_analysis}
\end{figure}

In the survivor case, token selection was distributed relatively evenly across time, with 60m, 120m, and 240m tokens retained throughout the sequence, as shown in Figure~\ref{fig:case_analysis}(a). Consistently, the mean gate remained high, whereas the mean staleness stayed low with little fluctuation, as shown in Figure~\ref{fig:case_analysis}(b). This pattern suggests that, under a stable observation flow, recent information was continuously regarded as reliable, allowing the model to use temporal context broadly rather than concentrating on a small number of intervals.

In contrast, in the non-survivor case, token selection became increasingly concentrated toward the later period, with shorter-resolution tokens, particularly 60m and 120m, being preferentially retained, as shown in Figure~\ref{fig:case_analysis}(c). Figure~\ref{fig:case_analysis}(d) further shows that mean staleness varied more substantially, whereas mean gate was generally lower than in the survivor case, although it remained relatively sustained when recent observations were available. This suggests that, under increasing observation irregularity, the model selectively retained information according to both recency and reliability. In particular, the concentration of short-resolution tokens in the later period is consistent with clinical decision-making, where abrupt recent physiological changes often play a central role in assessing mortality risk during rapid deterioration \cite{ref27,ref28}.

Overall, these cases show that ReTAMamba reflects cohort-level differences in observation patterns at the individual-patient level through gate values, staleness, and multi-scale token selection. This supports the ability of the proposed framework to capture short-term acute signs in irregular clinical time series.

\subsection{Multi-scale Effects}

This section analyzes the effect of multi-scale temporal configurations on model performance using different combinations of temporal scales. The single-scale setting was fixed at 60 minutes, reflecting that vital signs in the ICU are generally documented on an hourly basis~\cite{ref29}. Multi-scale settings were then constructed by progressively extending this base configuration to \{60, 120\}, \{60, 120, 240\}, and \{60, 120, 240, 480\}.

As shown in Table~\ref{tab:multiscale_effects}, multi-scale configurations generally outperformed the single-scale setting across all datasets. In particular, the combination of 60, 120, and 240 achieved the best AUROC and AUPRC on MIMIC-IV, eICU, and PhysioNet 2012, indicating that jointly modeling short-term changes and longer-term trends is effective for clinical time-series prediction. In contrast, adding the extended 480-minute scale did not yield further gains and even degraded some metrics. This suggests that overly coarse temporal resolution may oversmooth short-term changes while increasing information redundancy, reducing representational efficiency even with token routing.

Overall, these results show that multi-scale aggregation is important for capturing diverse temporal patterns in irregular clinical time series, and that the combination of short-, medium-, and long-range scales provides the best trade-off between predictive performance and efficiency.

\section{Conclusion}

This paper proposes ReTAMamba for effective modeling of clinical time series with irregular sampling and missingness. ReTAMamba reconstructs clinical time series as time-variable token sequences, models observation reliability from missingness and elapsed time, and combines multi-scale aggregation with chronological token organization and budgeted routing. Through this design, it preserves rich temporal context while efficiently controlling sequence length.

Experiments on three clinical time-series benchmarks, MIMIC-IV, eICU, and PhysioNet 2012, showed that ReTAMamba consistently outperforms existing baselines in both AUROC and AUPRC under a unified in-hospital mortality prediction setting, with statistically significant AUPRC improvements over the strongest competing models on all three datasets. The ablation study further confirmed that the Reliability Gate, Stats Augment, Chronological Weaving, and Token Router each contribute to the performance gains. Efficiency and temporal pattern analyses also showed that the proposed model provides a practical trade-off among predictive performance, computational cost, and the effective use of observation patterns.

These findings suggest that effective prediction from irregular clinical time series requires modeling not only measured values themselves but also observation structure, observation reliability, information recency, multi-scale temporal context, and budgeted sequence compression. Although this study focused on mortality prediction under a unified early-risk assessment setting, the same framework could be extended to other clinical tasks such as decompensation, length of stay, and readmission, and could further be examined for irregular time-series data in other domains.

\appendix
\renewcommand{\thetable}{A\arabic{table}}
\setcounter{table}{0}

\section{Additional Experimental Details}

\subsection{Final Hyperparameters}

Table~\ref{tab:final_hyperparameters} summarizes the final hyperparameter settings used for ReTAMamba.

\begin{table}[h]
\centering
\caption{Final hyperparameter settings for ReTAMamba.}
\label{tab:final_hyperparameters}
\footnotesize
\begin{tabular}{lll}
\toprule
Hyperparameter & Search Range & Final Selection \\
\midrule
lr & $[3\times10^{-5},\,3\times10^{-3}]$ (log scale) & $5.904\times10^{-4}$ \\
weight decay & $[10^{-6},\,10^{-2}]$ (log scale) & $1.709\times10^{-6}$ \\
dropout & $[0.00,\,0.30]$ & $0.032$ \\
grad clip & $\{0.5,\,1.0,\,2.0,\,5.0\}$ & $0.5$ \\
$d_{\text{model}}$ & $\{64,\,96,\,128,\,160,\,192,\,256\}$ & $96$ \\
$n_{\text{layers}}$ & $[2,\,8]$ & $5$ \\
$d_{\text{state}}$ & $\{16,\,32,\,64,\,96\}$ & $96$ \\
$d_{\text{conv}}$ & $\{2,\,3,\,4\}$ & $2$ \\
expand & $\{1,\,2,\,3,\,4\}$ & $3$ \\
init decay logit & $[-6.0,\,-1.0]$ & $-2.717$ \\
$\lambda_{\min}$ & $[10^{-4},\,10^{-2}]$ (log scale) & $4.289\times10^{-4}$ \\
token budget $k$ & $\{8,\,16,\,32,\,48\}$ & $32$ \\
temporal scales & 4 candidate sets & $\{60,120,240\}$ \\
pooling & \{attn, mean, last\} & last \\
\bottomrule
\end{tabular}
\end{table}

\subsection{Token Budget Sensitivity}

To examine the effect of routing budget, we conducted a supplementary sensitivity analysis by varying $k$ while keeping the other model settings fixed. For reference, Table~\ref{tab:pre_routing_tokens} reports the mean and standard deviation of the number of valid multi-scale tokens before token routing.

\begin{table}[h]
\centering
\caption{Pre-routing token statistics across datasets.}
\label{tab:pre_routing_tokens}
\footnotesize
\begin{tabular}{lcc}
\toprule
Dataset & Mean tokens & Std tokens \\
\midrule
MIMIC-IV       & $46.70$ & $3.31$ \\
eICU          & $46.72$ & $5.09$ \\
PhysioNet 2012 & $44.60$ & $6.71$ \\
\bottomrule
\end{tabular}
\end{table}

\begin{table}[h]
\centering
\caption{Token budget sensitivity analysis.}
\label{tab:k_budget_sensitivity}
\footnotesize
\begin{tabular}{lcccccc}
\toprule
\multirow{2}{*}{\raisebox{-0.9ex}{$k$ (budget)}} & \multicolumn{2}{c}{MIMIC-IV} & \multicolumn{2}{c}{eICU} & \multicolumn{2}{c}{PhysioNet 2012} \\
\cmidrule(lr){2-3}\cmidrule(lr){4-5}\cmidrule(lr){6-7}
& AUROC & AUPRC & AUROC & AUPRC & AUROC & AUPRC \\
\midrule
router off  & 0.858 & 0.538 & 0.780 & 0.361 & 0.841 & 0.512 \\
8           & 0.832 & 0.518 & 0.776 & 0.358 & 0.831 & 0.496 \\
16          & 0.849 & 0.531 & 0.783 & 0.364 & 0.838 & 0.507 \\
\textbf{32} & \textbf{0.861} & \textbf{0.548} & \textbf{0.793} & \textbf{0.384} & \textbf{0.847} & \textbf{0.520} \\
48          & 0.855 & 0.536 & 0.781 & 0.362 & 0.842 & 0.514 \\
\bottomrule
\end{tabular}
\end{table}

Table~\ref{tab:k_budget_sensitivity} summarizes predictive performance under different routing budgets. Overly small budgets degraded performance across datasets, while performance improved at intermediate budgets. The best AUROC and AUPRC were consistently achieved at $k=32$, supporting the use of a moderate token budget as an effective trade-off between predictive performance and sequence compression.

\subsection{Calibration Results}

To complement the discrimination results reported in the main text, we additionally evaluated probability calibration on the eICU dataset. We chose eICU for this supplementary analysis because it is a multi-center benchmark with substantial heterogeneity in observation patterns, making calibration assessment particularly informative under diverse clinical settings. We report the Brier score and expected calibration error (ECE), where the Brier score reflects the overall accuracy of predicted probabilities and ECE measures the agreement between predicted confidence and empirical outcomes across probability bins. Lower values indicate better calibration.

\begin{table}[h]
\centering
\caption{Calibration results on the eICU dataset. Lower is better for both metrics.}
\label{tab:calibration_results}
\small
\begin{tabular}{lcc}
\toprule
Model & Brier $\downarrow$ & ECE $\downarrow$ \\
\midrule
XGBoost     & $0.233 \pm 0.000$ & $0.351 \pm 0.000$ \\
LSTM        & $0.175 \pm 0.015$ & $0.248 \pm 0.034$ \\
Transformer & $0.183 \pm 0.015$ & $0.257 \pm 0.023$ \\
Mamba       & $0.181 \pm 0.008$ & $0.262 \pm 0.008$ \\
GRU-D       & $0.182 \pm 0.011$ & $0.254 \pm 0.020$ \\
IP-Nets     & $0.187 \pm 0.018$ & $0.274 \pm 0.035$ \\
mTAN        & $0.183 \pm 0.048$ & $0.288 \pm 0.075$ \\
SeFT        & $0.202 \pm 0.012$ & $0.300 \pm 0.018$ \\
Raindrop    & $0.215 \pm 0.043$ & $0.314 \pm 0.063$ \\
Warpformer  & $0.172 \pm 0.024$ & $0.238 \pm 0.045$ \\
\textbf{ReTAMamba }  & $\mathbf{0.162 \pm 0.011}$ & $\mathbf{0.221 \pm 0.018}$ \\
\bottomrule
\end{tabular}
\end{table}

Table~\ref{tab:calibration_results} reports the calibration results on eICU. ReTAMamba achieved the lowest Brier score and the lowest ECE among all compared models, indicating the best calibration quality in this supplementary evaluation. Compared with the average of the baseline models, ReTAMamba reduced the Brier score by 15.3\% and the ECE by 20.7\%. Compared with the strongest baseline, Warpformer, it further reduced the Brier score from $0.172$ to $0.162$ and the ECE from $0.238$ to $0.221$. These results suggest that the proposed model not only improves discriminative performance but also provides more reliable probability estimates on the eICU benchmark.

\bibliographystyle{ACM-Reference-Format}
\bibliography{references}
\end{document}